%% file: main.tex
\documentclass[10pt,twocolumn,letterpaper]{article}

 \usepackage[pagenumbers]{cvpr} % To force page numbers, e.g. for an arXiv version

\input{preamble}
\definecolor{cvprblue}{rgb}{0.21,0.49,0.74}
\usepackage[pagebackref,breaklinks,colorlinks,allcolors=cvprblue]{hyperref}
\usepackage{multirow}

\usepackage{pifont,xcolor,booktabs}
\newcommand{\cmark}{\textcolor{green!60!black}{\ding{51}}}
\newcommand{\xmark}{\textcolor{red!65!black}{\ding{55}}}
\usepackage[most]{tcolorbox}
\definecolor{ecoGreen}{RGB}{46,139,87} % 主色调: 适度绿色 (类似sea green)

\tcbset{
  enhanced,
  breakable,
  colback=green!3,              % 背景浅绿
  colframe=ecoGreen,            % 边框深绿
  boxrule=0.5pt,
  arc=2mm,
  left=6pt, right=6pt, top=6pt, bottom=6pt,
  fonttitle=\bfseries,
  coltitle=black,
  title style={left color=ecoGreen!20,right color=white},
}

\newtcolorbox{toolbox}[1][]{
  title=\textbf{#1},
  colback=green!2,
  colframe=ecoGreen!80!black,
  boxrule=0.5pt,
  arc=2mm,
  top=5pt, bottom=5pt, left=6pt, right=6pt,
  fonttitle=\bfseries,
  coltitle=black,
  enhanced,
}

\usepackage[ruled,vlined]{algorithm2e} % <-- 注释掉或删除

\SetKwInput{KwInput}{Input}
\SetKwInput{KwOutput}{Output}
\def\paperID{3489} % *** Enter the Paper ID here
\def\confName{CVPR}
\def\confYear{2026}

\title{
    GreenPlanner: Practical Floorplan Layout Generation via an Energy-Aware and Function-Feasible Generative Framework
}

\author{
  \large
  Pengyu Zeng\thanks{Equal contribution.}, 
  Yuqin Dai\footnotemark[\value{footnote}],  % 复用前一个标记（*）
  Jun Yin\footnotemark[\value{footnote}],    % 复用前一个标记（*）
  Jing Zhong, 
  Ziyang Han, 
  Chaoyang Shi, \\
  \large
  ZhanXiang Jin, 
  Maowei Jiang, 
  Yuxing Han, 
  Shuai Lu\thanks{Corresponding author. E-mail: shuai.lu@sz.tsinghua.edu.cn}\\
  \large Tsinghua University
}

\begin{document}
\maketitle
\input{sec/0_abstract}

\input{sec/1_intro}

\input{sec/2_relatedwork}

\input{sec/3_method}
\input{sec/4_experiments}

\input{sec/5_conclusion}
{
    \small
    \bibliographystyle{ieeenat_fullname}
    \bibliography{main}
}


\end{document}

%% file: sec/0_abstract.tex
\begin{abstract}
Building design directly affects human well-being and carbon emissions, yet generating spatial-functional and energy-compliant floorplans remains manual, costly, and non-scalable.
Existing methods produce visually plausible layouts but frequently violate key constraints, yielding invalid results due to the absence of automated evaluation.
We present \textbf{GreenPlanner}, an energy- and functionality-aware generative framework that unifies design evaluation and generation.
It consists of a labeled \textbf{Design Feasibility Dataset} for learning constraint priors; a fast \textbf{Practical Design Evaluator (PDE)} for predicting energy performance and spatial-functional validity; a \textbf{Green Plan Dataset (GreenPD)} derived from PDE-guided filtering to pair user requirements with regulation-compliant layouts; and a \textbf{GreenFlow} generator trained on GreenPD with PDE feedback for controllable, regulation-aware generation.
Experiments show that GreenPlanner accelerates evaluation by over $10^{5}\times$ with $>$99\% accuracy, eliminates invalid samples, and boosts design efficiency by 87\% over professional architects.
\textbf{Code and data will be available.}
\end{abstract}

%% file: sec/1_intro.tex
\section{Introduction}

\begin{figure}[t]
\centering
\includegraphics[width=1\linewidth]{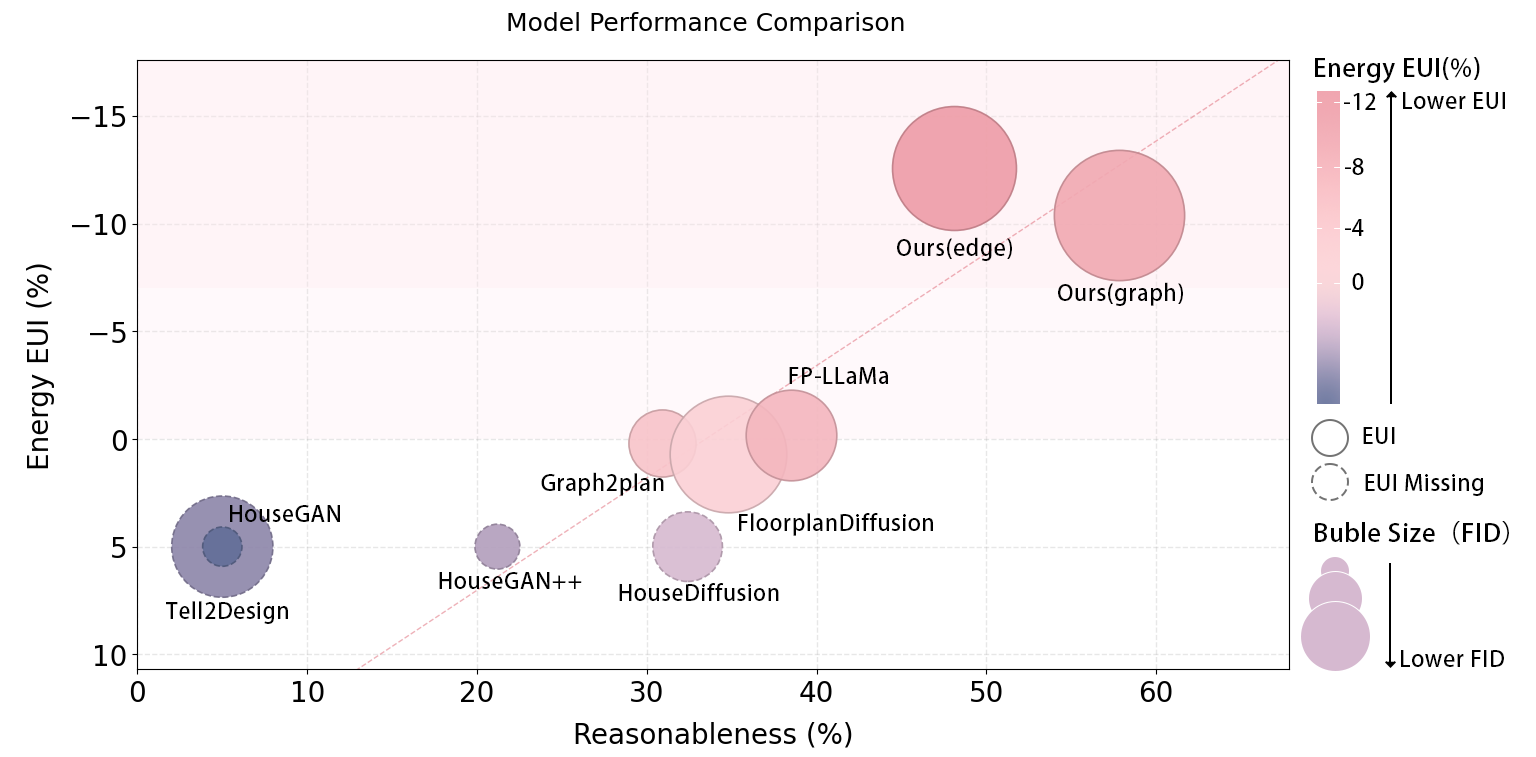}
\caption{Comparison of model performance in terms of reasonableness (higher is better), energy EUI (lower is better), and FID (bubble size, larger is better). Our method achieves higher spatial plausibility, significantly lower energy consumption, and better generation quality, surpassing existing approaches overall.}
\label{fig:overview}
\end{figure}

Humans spend nearly 90\% of their time indoors~\cite{dales2008quality}, and the built environment contributes a substantial share of global carbon emissions~\cite{liang2023decarbonization, zhang2023leveraging, zhang2024globus}. Thus, building design sits at the nexus of daily life and sustainability.
Traditional design relies heavily on expert knowledge and extensive manual iteration~\cite{baduge2022artificial,eberhardt2022building,himeur2023ai}, often taking days to months to reach viable solutions. 
Recent advances in generative models on controllable~\cite{chen2025comprehensive,wang2025diffusion,li2025comprehensive} and multimodal generation~\cite{gpt4o, gemini, xue2025human} pave the way for automated building design generation.

To enable automation, prior works have employed conditioned generators~\cite{rplan,Graph2Plan,HouseGAN,housegan++,FloorplanDiffusion,tell2design,housediffusion,zeng2025automated,zeng2025card,zeng2025comprehensive,zeng2025unified} that translate coarse user requirements into concrete floorplan layouts. However, existing methods rely solely on user-specified conditions and thus often violate unspecified real-world design constraints. 
This neglect results in over 60\% of samples in existing datasets~\cite{rplan} being practically unusable due to excessive energy use and flawed spatial-functional design. 
Some methods~\cite{housegan++,housediffusion} even fail to generate essential structural components such as windows, leading to unrealistic energy evaluations. 
Therefore, enforcing design constraints in generative models remains a major bottleneck for practical deployment.

Our observation is that the core bottleneck is the lack of an effective, practical evaluation strategy. Existing design pipelines~\cite{li2024revision, lee2023simplified, regenwetter2023beyond, chen2025gdt, zeng2025ai} still rely on manual \textit{design–evaluate–revise} cycles and slow energy simulations, making it difficult to assess energy performance and flawed spatial-functional rationality efficiently.
Such costly and non-scalable processes hinder the creation of high-quality annotated datasets and prevent models from learning practical designs. 
To overcome this, an automated framework is needed to efficiently evaluate and optimize generation under design constraints.

To tackle these challenges, we propose GreenPlanner, a generative framework for practical residential design that ensures both sustainability and spatial-functional rationality.
GreenPlanner integrates four key components to unify evaluation and generation.
First, we construct a \textbf{Design Feasibility Dataset (DesignFD)}, built from energy and spatial-functional evaluations with expert verification, providing quantitative and qualitative feasibility labels for model training.
Second, we develop a \textbf{Practical Design Evaluator (PDE)}, trained on DesignFD to enable instant estimation of energy performance and spatial-functional validity.
Third, we build a \textbf{Green Plan Dataset (GreenPD)}, derived from DesignFD by resampling and correcting non-compliant layouts under PDE guidance, pairing design requirements with regulation-compliant floorplans.
Finally, we introduce \textbf{GreenFlow}, a guided flow-based generator trained on GreenPD and refined through PDE feedback, enabling controllable, regulation-aware layout generation.
Together, these components form an end-to-end energy-aware framework that achieves clearly superior performance (see Fig.~\ref{fig:overview}).

In summary, our contributions are:
\begin{itemize}
    \item We introduce GreenPlanner, an energy-efficient residential design framework that seamlessly integrates evaluation and generation.  
    \item We construct DesignFD, a labeled feasibility dataset, and GreenPD, a demand–plan dataset ensuring regulation-compliant floorplans.  
    \item We develop PDE and GreenFlow, enabling fast evaluation and controllable generation under real-world constraints.  
    \item Extensive experiments show GreenPlanner’s effectiveness: PDE achieves over $\mathbf{10^{5}\times}$ speed-up (only $\sim$7.3\,ms per 100 cases) with $\mathbf{99\%}$ accuracy; GreenPD eliminates invalid samples ($\mathbf{60\%\!\to\!0\%}$), and GreenFlow boosts design efficiency by $\mathbf{87\%}$ vs. professional architects.
\end{itemize}

%% file: sec/2_relatedwork.tex
\section{Related Work}
\subsection{Generative Models}
Generative models have achieved remarkable progress across diverse content-creation domains~\citep{hu2025simulating, xue2025human, liu2025large, ding2025understanding, alvur2025potential,jiang2023fecam,zeng2022muformer,zeng2023seformer,xue2025human,dai2025harmonious,dai2025mindaligner,dai2025tcdiff++}.
Variational autoencoders~\citep{kingma2013auto} encode compact latent representations but struggle to capture complex high-dimensional distributions.
Generative adversarial networks~\citep{goodfellow2020generative} produce realistic results yet lack explicit likelihoods and suffer from training instability~\citep{sengar2025generative}.
Diffusion models~\citep{ho2020denoising, song2020denoising} generate high-quality and diverse samples, but their iterative denoising process is computationally expensive~\citep{croitoru2023diffusion, chang2025design}.
In contrast, flow-based models~\citep{bond2021deep, kobyzev2020normalizing, papamakarios2021normalizing} enable exact likelihood estimation and efficient sampling, offering an optimal balance of speed, fidelity, and training stability. 
Building on these strengths, GreenPlanner employs a flow-based generator, GreenFlow, for energy-efficient and functionally valid design generation.

\subsection{Floorplan Layout Generation}
Residential layout generation has evolved from topology- and geometry-driven synthesis to multimodal approaches~\cite{zeng2025mred}. 
Representative topology-based methods include HouseGAN~\cite{HouseGAN,housegan++}, Graph2Plan~\cite{Graph2Plan}, and HouseDiffusion~\cite{housediffusion}, which generate layouts from graph or block representations. 
More recent diffusion- and language-conditioned frameworks, such as FloorplanDiffusion~\cite{zeng2024residential}, Tell2Design~\cite{tell2design}, LayoutGPT~\cite{layoutgpt}, and Holodeck~\cite{holodeck}, improve diversity and controllability. 
However, many existing methods still produce structurally incomplete layouts~\cite{yin2025floorplanmae,yin2025floorplan,yin2025drag2build++,yin2025archiset}, for example, missing windows~\cite{housediffusion,housegan++} or lacking walls and doors~\cite{tell2design,HouseGAN}, which hinders energy estimation and practical deployment. 
GreenPlanner addresses these issues by integrating spatial and energy feasibility into the generation process, yielding layouts that are both functionally valid and energy-efficient.

%% file: sec/3_method.tex
\section{Method}

\begin{figure*}[!t]
\centering
\includegraphics[width=0.97\linewidth]{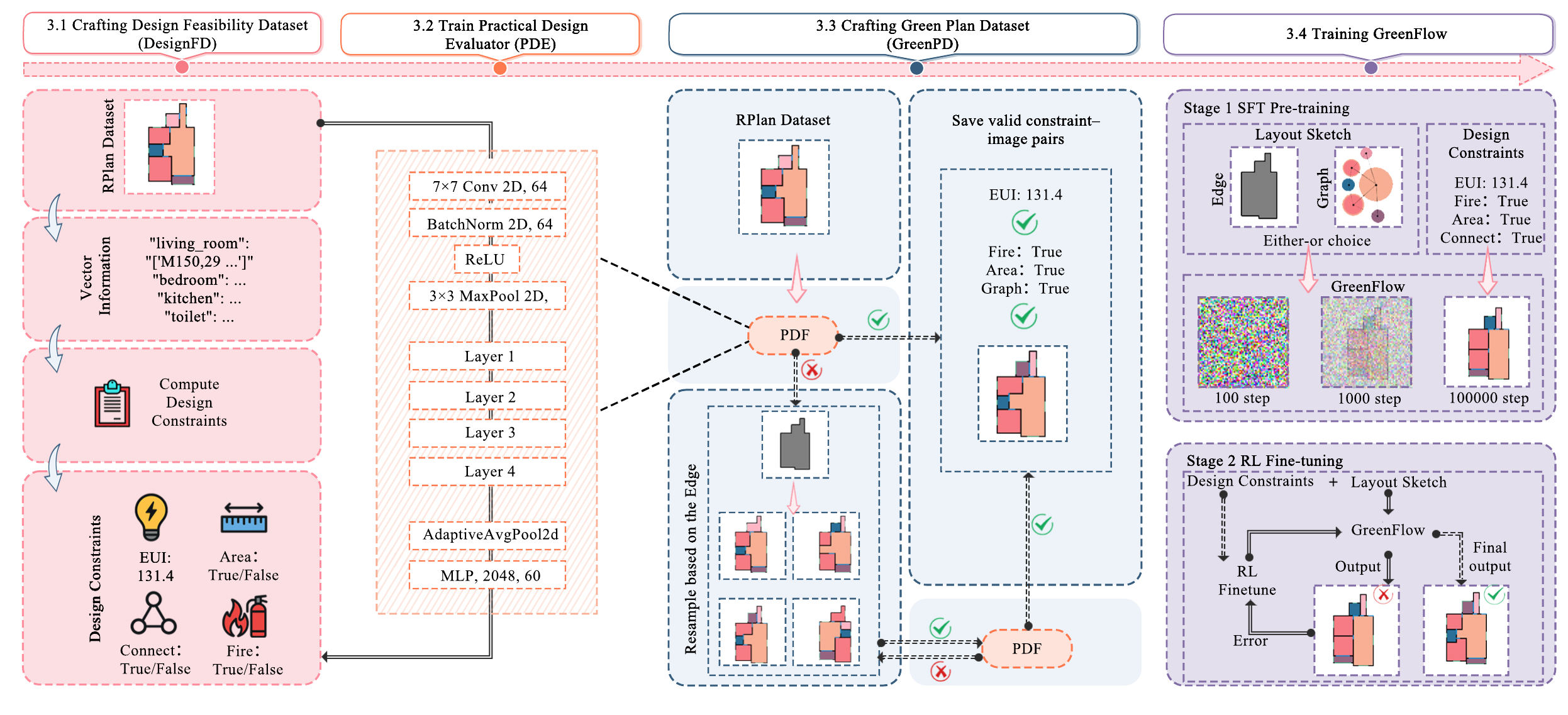}
\caption{Overview of our \textbf{GreenPlanner}, an energy- and functional-aware generative framework for residential layout design.
It consists of four key components: (1) Design Feasibility Dataset \textbf{(DesignFD)}, a dataset derived from RPLAN through quantitative energy and functional evaluations; (2) the Practical Design Evaluator \textbf{(PDE)}, a convolutional network trained on DesignFD to predict design metrics rapidly; (3) Green Plan Dataset \textbf{(GreenPD)}, a fully regulation-compliant demand–plan dataset refined via PDE-guided resampling; and (4) the \textbf{GreenFlow} generator, trained and fine-tuned under PDE feedback for controllable, regulation-aware floorplan generation.
Together, these components form an end-to-end pipeline from data construction to constraint-driven generative design.}
\label{fig:framework}
\end{figure*}

We propose \textbf{GreenPlanner}, an energy- and functionality-aware generative framework for residential design (Fig.~\ref{fig:framework}).
It integrates four components:
(1) \textbf{Design Feasibility Dataset (DesignFD)}, a dataset built from RPLAN~\cite{rplan} via expert-verified energy and functional evaluations;
(2) \textbf{Practical Design Evaluator (PDE)}, a predictor trained on DesignFD for fast simulation of energy and functional feasibility;
(3) \textbf{Green Plan Dataset (GreenPD)}, a PDE-guided dataset pairing user requirements with regulation-compliant layouts; and
(4) \textbf{GreenFlow}, a generator trained on GreenPD with PDE feedback for controllable, constraint-grounded layout generation.
We next describe each component of the framework in detail.

\subsection{Design Feasibility Dataset}
\paragraph{Quantitative and Qualitative Design Constraints.}
\label{sec:constrain_metrics}
We construct a Design Feasibility Dataset \textbf{(DesignFD)}, derived from RPLAN~\cite{rplan} with expert-verified energy and functional evaluations providing quantitative feasibility labels.
Specifically, DesignFD defines four key constraints: 
\textbf{(a)} \textbf{\textit{Energy Use Intensity (EUI)}}: calculated via annual energy simulations in EnergyPlus\footnote{U.S. Department of Energy, \textit{EnergyPlus™ Version 24.2.0 Engineering Reference}.}, where a lower EUI indicates higher energy efficiency; 
\textbf{(b)} \textbf{\textit{Fire-safety distance}}: the maximum walkable distance from the entrance to the farthest accessible point, required to be $\leq15$\,m under residential fire codes\footnote{GB 50016-2014: \textit{Code for Fire Protection Design of Buildings}.}; 
\textbf{(c)} \textbf{\textit{Floor area}}: derived from pixel-to-scale conversion, expected to remain within a reasonable range following standard design practice\footnote{GB 50096-2011: \textit{Design Code for Residential Buildings}.}; 
and \textbf{(d)} \textbf{\textit{Functional connectivity}}: assessed through graph-based analysis of room adjacencies to ensure rational circulation and usability\footnote{According to GB 50368-2005: \textit{Code for Design of Residential Buildings}, and GB 50352-2019: \textit{General Code for Civil Building Design}, residential layouts should maintain reasonable spatial organization and convenient circulation.}

\paragraph{Crafting DesignFD.}
DesignFD pairs each floorplan image with its corresponding energy and functional metrics, providing data support for quantitative feasibility modeling.  
We use automated analysis assisted by expert verification to compute four key indicators:  
\textbf{(a)} \textit{Energy Use Intensity (EUI)}: calculated through annual EnergyPlus simulations, indicating energy consumption per unit area;  
\textbf{(b)} \textit{Fire-safety distance}: measured via a walkable-grid search of the maximum accessible path (detailed in Appendix.~A1);  
\textbf{(c)} \textit{Floor area}: derived from pixel-to-scale conversion; and  
\textbf{(d)} \textit{Functional connectivity}: evaluated by comparing room-adjacency graphs with residential design codes (detailed in Appendix.~A2).  
Each floorplan is rasterized and annotated with continuous metrics (e.g., EUI) and binary feasibility labels, forming paired data for PDE training.

\subsection{Practical Design Evaluator}

\paragraph{Model Design.}
The Practical Design Evaluator (\textbf{PDE}) predicts all DesignFD metrics directly from rasterized floorplans. 
Although transformer-based models~\cite{vaswani2017attention} capture global semantics, they often compromise local spatial coherence. 
Convolutional networks, by contrast, better preserve spatial topology and geometric bias. 
Therefore, we adopt a convolutional regression network (illustrated in Fig.~\ref{fig:framework} and detailed in Appendix.~A3) for accurate and stable prediction of physical feasibility metrics.

\begin{table*}[!t]
\centering
\caption{
Comparison of floorplan datasets. 
The first two columns (\textit{Annotations}) list available labeled data, and the last three (\textit{Applicable Tasks}) indicate the design capabilities supported by each dataset. 
\textit{Sketch} provides conceptual layouts, while \textit{Design Constraints} include expert-assessed metrics such as EUI, fire distance, floor area, and connectivity. 
\textit{Plan Generation}, \textit{Regulation Evaluation}, and \textit{Compliance Generation} represent progressively more constrained and goal-oriented design tasks. 
RPLAN focuses on sketch-based generation, 
DesignFD enables constraint-guided evaluation, 
and GreenPD advances to fully regulated, compliance-driven generation.
}
\resizebox{\linewidth}{!}{
\begin{tabular}{l|cc|ccc}
\toprule
\textbf{Dataset} & \multicolumn{2}{c|}{\textbf{Annotations}} & \multicolumn{3}{c}{\textbf{Applicable Tasks}} \\
\cmidrule(lr){2-3} \cmidrule(lr){4-6}
& \textbf{Sketch} & \textbf{Design Constraints (\S\ref{sec:constrain_metrics})} 
& \textbf{Plan Generation} & \textbf{Regulation Evaluation} & \textbf{Compliance Generation} \\
\midrule
RPLAN~\cite{rplan} 
& \cmark 
& \xmark 
& \cmark 
& \xmark 
& \xmark\ (low compliance) \\
DesignFD (ours) 
& \xmark 
& \cmark\ (EUI, fire distance, floor area, connectivity) 
& \xmark 
& \cmark 
& \xmark\ \\
GreenPD (ours) 
& \cmark 
& \cmark 
& \cmark 
& \xmark 
& \cmark\ (100\%) \\
\bottomrule
\end{tabular}}
\label{tab:dataset_comparison}
\end{table*}

\paragraph{Training Procedure.}
\label{sec:pde_trianing}
PDE is trained using the Mean Squared Error (MSE) loss over all quantitative and qualitative targets in DesignFD, formulated as:
\begin{equation}
\mathcal{L}_{\text{MSE}} = \frac{1}{N} \sum_{i=1}^{N} \| \hat{\mathbf{y}}_i - \mathbf{y}_i \|_2^2,
\end{equation}
where $\hat{\mathbf{y}}_i$ and $\mathbf{y}_i$ denote the predicted and ground-truth metric vectors for the $i$-th sample, respectively. 
Each target vector is defined as 
\begin{equation}
\mathbf{y}_i = [\mathbf{E}_i, F_i, A_i, G_i] \in \mathbb{R}^{63},
\end{equation}
where $\mathbf{E}_i \!\in\! \mathbb{R}^{60}$ represents the \textit{energy-use intensity} profile derived from annual energy simulations across 60 thermal and equipment parameters;  
$F_i \!\in\! \mathbb{R}$ denotes the \textit{fire-safety distance} measuring the maximum walkable path within code limits;  
$A_i \!\in\! \mathbb{R}$ indicates the \textit{floor area} reflecting geometric reasonableness;  
and $G_i \!\in\! \mathbb{R}$ quantifies the \textit{functional connectivity} based on room-adjacency graph consistency.  
Together, these 63 features comprehensively encode both energy efficiency and functional feasibility within DesignFD. 
As validated in Appendix.~A3, PDE achieves $R^2 > 0.99$ accuracy against full EnergyPlus simulations while operating orders of magnitude faster, demonstrating strong practicality for downstream generative design tasks.

\subsection{Green Plan Dataset}

\paragraph{Demand–Plan Pairing.}
Green Plan Dataset (GreenPD) extends DesignFD into a paired dataset linking user requirements with fully compliant residential layouts.  
Each sample is a pair $(\mathbf{d}_i, \mathbf{x}_i)$, where $\mathbf{d}_i$ denotes the user demand and $\mathbf{x}_i$ the corresponding floorplan.  
Specifically, the demand $\mathbf{d}_i$ is composed of two parts:
$\mathbf{d}_i = (\mathbf{g}_i, \mathbf{t}_i)$, where  
(1) $\mathbf{g}_i$ is a \textit{layout sketch}: an image representation that encodes either the room adjacency graph (inter-room connectivity) or the exterior boundary; and  
(2) $\mathbf{t}_i$ is a \textit{design constraint}: a textual description represented as  
$\mathbf{t}_i = [\mathbf{E}_i, F_i, A_i, G_i]$,  
specifying the target energy-use profile, fire-safety distance, total floor area, and functional connectivity validity (defined in Sec.~\ref{sec:pde_trianing}).  
This unified formulation allows GreenFlow to learn conditional generation driven jointly by spatial layout priors ($\mathbf{g}_i$) and design constraints ($\mathbf{t}_i$), enabling design generation that adheres to regulatory requirements.

\paragraph{Dataset Construction.}
GreenPD is derived from DesignFD through PDE-guided refinement to ensure full physical compliance.
Among the 71{,}125 layouts in RPLAN~\cite{rplan}, 26{,}600 valid samples are directly preserved, while the remaining 44{,}525 undergo iterative correction.
For each non-compliant footprint, FloorplanDiffusion~\cite{zeng2024residential} generates candidate layouts conditioned on the same boundary or room configuration.
PDE then evaluates their energy and functional metrics, retaining only those that meet all feasibility criteria.
This resampling process is repeated until every case satisfies the DesignFD constraints, yielding a fully regulation-compliant dataset for GreenFlow training.
A comparative summary in Tab.~\ref{tab:dataset_comparison} highlights the progressive evolution from RPLAN to DesignFD and GreenPD, emphasizing the dataset contributions that enhance regulatory awareness and compliance through expert-labeled constraints and high-quality annotations.

\subsection{GreenFlow}
\label{sec:greenflow}
GreenFlow adopts a two-stage learning paradigm for energy- and functional-aware layout generation:
(1) supervised pre-training with paired textual–topological constraints, and
(2) PDE-guided reinforcement fine-tuning.
This allows the model to first learn realistic layout priors and then align them with functional constraints.

\paragraph{Stage 1: Supervised Pre-training.}
As shown in Algorithm~\ref{alg:greenflow_stage1}, GreenFlow is first trained to reconstruct floorplans from paired textual–topological demands.  
Each input demand $\mathbf{d}_i = (\mathbf{g}_i, \mathbf{t}_i)$ consists of a layout sketch $\mathbf{g}_i$ and a design constraint $\mathbf{t}_i$, which jointly specify requirements such as energy-use level, fire-safety distance, total floor area, and room connectivity, as illustrated in Fig.~\ref{fig:framework}.  
The textual component $\mathbf{t}_i$ is encoded via CLIP~\cite{clip}, while the topological component $\mathbf{g}_i$ is transformed into a graph embedding.  
These embeddings are concatenated to condition the generator $G_\theta$, which is optimized using an $L_1$ loss and a perceptual loss~\cite{johnson2016perceptual} against the ground-truth layout.  
This stage enables GreenFlow to learn the relationship between user demands and layouts.

\begin{algorithm}[t]
\small
\setlength{\algomargin}{0.8em}
\DontPrintSemicolon
\SetAlgoNlRelativeSize{-1}
\SetKwInput{KwIn}{Input}
\SetKwInput{KwOut}{Output}
\SetKwComment{Comment}{$\triangleright$\ }{}

\caption{Stage~1: GreenFlow Pre-training}
\label{alg:greenflow_stage1}

\KwIn{Paired data $\{(\mathbf{x}_i,\mathbf{d}_i)\}$, where 
$\mathbf{x}_i$ is a floorplan image and 
$\mathbf{d}_i=(\mathbf{g}_i,\mathbf{t}_i)$ is a user demand consisting of 
a layout sketch $\mathbf{g}_i$ (topology or boundary) and 
a design constraint $\mathbf{t}_i$ (e.g., energy-use profile, fire-safety distance, total area, connectivity);
text encoder $\mathrm{Enc}_{\text{text}}$; 
graph encoder $\mathrm{Enc}_{\text{graph}}$; 
generator $G_\theta$; 
loss weight $\lambda_{\text{perc}}$.}
\KwOut{Pretrained generator $G_\theta$.}

\ForEach{$(\mathbf{x}_i,\mathbf{d}_i)$}{
  $\mathbf{p}_i \leftarrow \mathrm{Enc}_{\text{text}}(\mathbf{t}_i)$; \ 
  $\mathbf{g}'_i \leftarrow \mathrm{Enc}_{\text{graph}}(\mathbf{g}_i)$; \ 
  $\mathbf{z}_i \leftarrow \mathrm{Concat}(\mathbf{p}_i,\mathbf{g}'_i)$; \ 
  $\hat{\mathbf{x}}_i \leftarrow G_\theta(\mathbf{z}_i)$; \ 
  $\mathcal{L}^{(i)} = \|\hat{\mathbf{x}}_i-\mathbf{x}_i\|_1 + 
  \lambda_{\text{perc}}\mathcal{L}_{\text{perc}}(\hat{\mathbf{x}}_i,\mathbf{x}_i)$; \ 
  $\theta \leftarrow \theta - \eta \nabla_\theta \mathcal{L}^{(i)}$.
}
\end{algorithm}

\paragraph{Stage 2: Reinforcement Fine-tuning.}
To enhance functional controllability, GreenFlow is refined through a gradient-based, model-driven reinforcement learning (RL) process under PDE supervision~\cite{schulman2017proximal,lehmann2024definitive}.  
The PDE acts as a differentiable environment that evaluates generated layouts and provides continuous feedback.  
Given a pretrained generator $G_\theta$ and the PDE evaluator $f_{\text{PDE}}$, a layout is generated from textual–topological conditions:
\begin{align}
\hat{\mathbf{x}} &= G_\theta(\mathbf{z}), \\
\mathbf{z} &= 
\mathrm{Concat}\!\big(
  \mathrm{Enc}_{\text{text}}(\mathbf{t}),\,
  \mathrm{Enc}_{\text{graph}}(\mathrm{Descriptor}(\mathbf{g}))
\big),
\end{align}
where $\mathrm{Descriptor}(\mathbf{g})$ encodes the functional attributes of the input topology graph, $\mathrm{Enc}{\text{text}}$ is a text encoder that embeds the design constraint $\mathbf{t}$, and $\mathrm{Enc}{\text{graph}}$ is a graph encoder that embeds the layout sketch $\mathbf{g}$.
The PDE predicts the corresponding performance metrics:
\begin{equation}
\hat{\mathbf{y}} = f_{\text{PDE}}(\hat{\mathbf{x}}),
\end{equation}
and computes a differentiable reward comparing $\hat{\mathbf{y}}$ with target constraints $\mathbf{y}^*$:
\begin{equation}
\mathcal{R} =
- \sum_{\mathbf{y}_k \in \mathbf{y}_{\text{spec}}}
\mathrm{MSE}(\hat{\mathbf{y}}_k, \mathbf{y}^*_k)
+ \lambda\,\mathcal{L}_{\text{rec}}(\hat{\mathbf{x}}, \mathbf{x}^*).
\end{equation}

Gradients of $\mathcal{R}$ are backpropagated through both the PDE and generator for parameter updates:
\begin{equation}
\theta \leftarrow \theta + \eta \nabla_\theta \mathcal{R},
\end{equation}
where $\eta$ is the learning rate.  
This model-based optimization aligns generation with energy and functional constraints while preserving geometric fidelity.  
Overall, GreenFlow integrates generation and evaluation within a differentiable pipeline, achieving stable, data-efficient fine-tuning and precise control over functional feasibility.

%% file: sec/4_experiments.tex
\section{Experiments}
\subsection{Experimental Setup}
\paragraph{Dataset and Baselines.}
We evaluate on the RPLAN~\cite{rplan} dataset and our GreenPD to test model generalization and compliance.
All baselines are retrained on the same split (70% train, 20% val, 10% test) for fairness.
We compare GreenFlow with representative methods across major paradigms:
GAN-based (HouseGAN~\cite{HouseGAN}, HouseGAN++~\cite{housegan++}),
diffusion-based (HouseDiffusion~\cite{housediffusion}, FloorplanDiffusion~\cite{zeng2024residential}),
and graph-to-image translation (Graph2Plan~\cite{Graph2Plan}).
Language-conditioned models (Tell2Design~\cite{tell2design}, FP-LLaMa~\cite{yin2025floorplan}) are also included.
All models are trained and tested under identical conditions.

\paragraph{Baseline Deficiency Handling.}
Some baselines produce incomplete layouts that hinder energy analysis.
\textit{HouseDiffusion} and \textit{HouseGAN++} often miss windows, inflating energy-use intensity (EUI).
We insert one to two windows per room for realistic correction.
Methods missing walls or doors (e.g., \textit{Tell2Design} or \textit{HouseGAN}) are excluded from energy evaluation to ensure fair comparison.

\paragraph{Metrics.}
We follow the evaluation protocol of previous layout generation works~\cite{HouseGAN, Graph2Plan, zeng2024residential} and extend it with additional metrics to jointly assess functional feasibility and energy efficiency. 
Five metrics are used for comprehensive evaluation:
(1) \textbf{Fréchet Inception Distance (FID)}~\cite{heusel2017gans}: computed with Inception-V3 embeddings to measure visual realism;  
(2) \textbf{Intersection over Union (IoU)}~\cite{everingham2010pascal}: quantifies geometric overlap between generated and ground-truth layouts;  
(3) \textbf{Functional Rationality (Rationality)}: the proportion of layouts meeting fire-safety, area, and connectivity requirements predicted by the \textit{PDE};  
(4) \textbf{Energy Use Intensity (EUI)}: estimated by \textit{PDE} to approximate annual energy demand per unit area.  
Both Rationality and EUI are computed by the \textit{PDE} and validated against full EnergyPlus simulations for physically consistent assessment.

\subsection{Implementation Details}
All experiments are implemented in PyTorch and trained on a single NVIDIA RTX A100 GPU. 
\textbf{Practical Design Evaluator~(PDE)} 
uses a ResNet-50 backbone~\cite{he2016deep} with standard residual blocks and batch normalization. 
Features are processed by global adaptive average pooling and mapped to a 63-dimensional output (3 for spatial-functional feasibility and 60 for energy-related metrics). 
The input resolution is $64\times64$. 
Training uses the Adam optimizer with an initial learning rate of $1\times10^{-4}$ and exponential decay. 
Early stopping (patience=3) is used to avoid overfitting. 
\textbf{GreenFlow} 
adopts a normalizing flow architecture optimized with the AdamW optimizer~\cite{adamw}. 
Stage~1 (supervised pre-training) runs for 200k iterations with batch size 16 and learning rate $1\times10^{-4}$, followed by Stage~2 (reinforcement fine-tuning) for 30k iterations with batch size 16 and learning rate $1\times10^{-5}$, using PDE feedback as reward signals.

\subsection{Compliance Analysis on GreenPD}

\begin{figure*}[!t]
\centering
\includegraphics[width=0.9\linewidth]{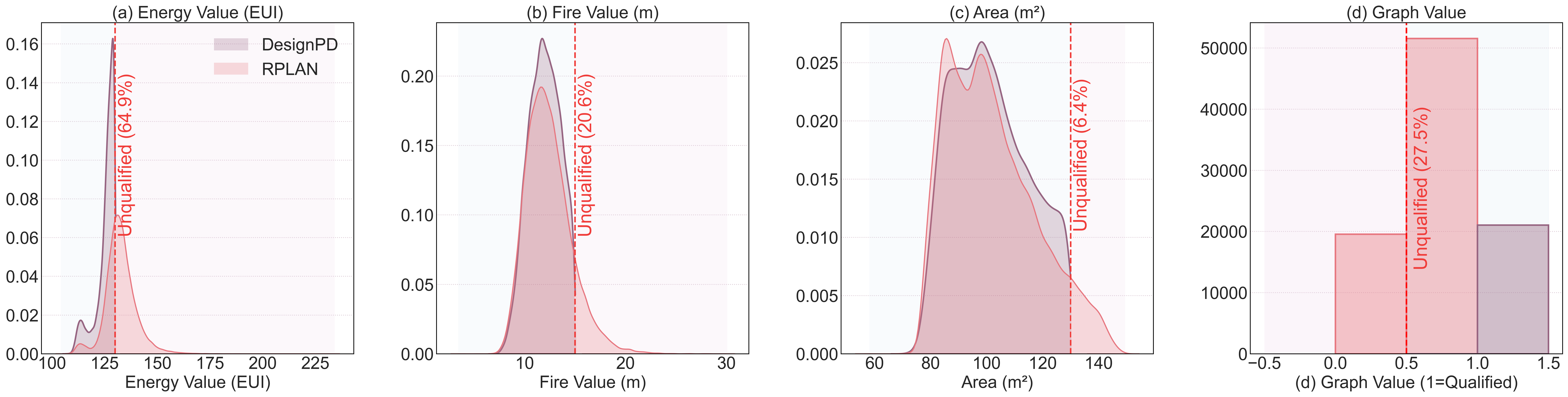}
\caption{Comparison between the original RPLAN dataset and our proposed GreenPD. 
While RPLAN contains numerous violations in energy, safety, and functional constraints, 
GreenPD achieves complete compliance with all physical design standards.}
\label{fig:data_distribution}
\end{figure*}

\paragraph{Existing datasets contain extensive infeasible data.}
Because the existing dataset~\cite{rplan} was constructed without enforcing building-code constraints and contains outdated or loosely labeled samples, our analysis reveals clear deficiencies in both energy and functional feasibility. 
As shown in Fig.~\ref{fig:data_distribution}, about 30.9\% of layouts exceed the energy-efficiency limit (EUI$>135$), 10.0\% violate the 15\,m fire-safety distance requirement, 9.7\% surpass the 130\,m$^2$ floor-area guideline, and 27.5\% fail to maintain functional connectivity between key rooms. 
Consequently, only 37.4\% of floorplans fully comply with the above national residential standards, indicating substantial non-compliance and potential bias that undermine model reliability.

\paragraph{Proposed GreenPD effectively corrects infeasible data.}
In contrast, our proposed GreenPD dataset (see Fig.~\ref{fig:data_distribution}) 
achieves \textit{\textbf{100\% compliance}} across all physical metrics by filtering and regenerating invalid samples. 
This ensures a clean, physically grounded dataset for model training, 
substantially improving both the functional feasibility and energy efficiency of generated layouts.

\subsection{PDE Analysis}
\begin{table}[t]
\centering
\caption{PDE performance summary. Runtime denotes total inference time over 100 runs.}
\label{tab:pde_summary}
\setlength{\tabcolsep}{6pt}
\renewcommand{\arraystretch}{1.05}
\small  
\begin{tabular}{lccc}
\toprule
\textbf{Method} & $R^2\uparrow$ & MSE$\downarrow$ & Runtime (total for 100 runs)$\downarrow$ \\
\midrule
PDE (Ours) & 0.99 & 0.46 & 7.3 ms \\
Baseline & -- & -- & $1.2{\times}10^6$ ms \\
\bottomrule
\end{tabular}
\end{table}
We evaluate the Practical Design Evaluator (PDE) against physics-based \textit{EnergyPlus} simulations. 
Tab.~\ref{tab:pde_summary} reports three standard regression metrics: 
$R^2$ (coefficient of determination) measures correlation with ground-truth variance, 
MSE (mean squared error) quantifies overall deviation, 
and runtime reflects total inference time over 100 runs. 
As shown, PDE achieves an $R^2$ of 0.99 and an MSE of 0.46, demonstrating near-perfect consistency with EnergyPlus while being over $1.6\times10^5$ times faster (7.3~ms vs.\ 1.2$\times$10$^6$~ms). 
These results confirm that PDE accurately predicts energy and spatial feasibility from rasterized floorplans, enabling rapid and scalable evaluation for generative design.

\subsection{Qualitative Comparison}

\begin{figure*}[t]
\centering
\includegraphics[width=0.9\linewidth]{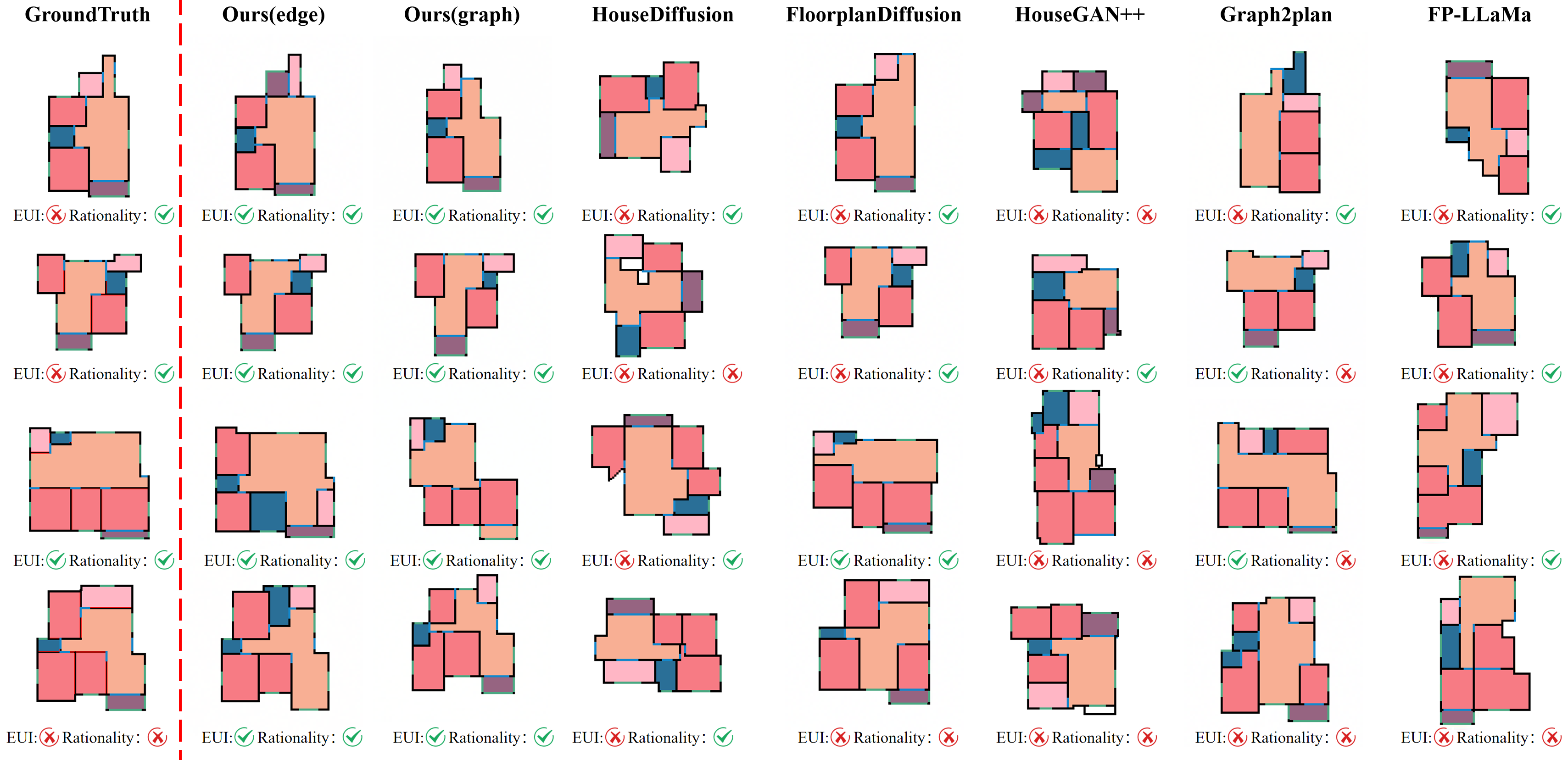}
\caption{
Qualitative comparison of layout generation results. Baseline models (e.g., HouseDiffusion, HouseGAN++) often fail to produce necessary openings (doors and windows), resulting in functional incomplete layouts and unrealistic energy performance. For fair evaluation, a small number of openings were randomly inserted into their outputs to enable energy assessment. Our GreenFlow generates layouts with coherent spatial organization and complete openings.
}
\label{fig:vis}
\end{figure*}

Most baselines omit openings, making their layouts both spatially disconnected and unsuitable for energy evaluation. To standardize simulation, we randomly add 1–2 openings to their outputs to meet the minimum boundary conditions required for computing EUI. However, as shown in Fig.~\ref{fig:vis}, these baselines still produce layouts with missing circulation paths, blocked room access, or inconsistent façade logic, indicating that opening placement is not merely a data annotation issue but a functional reasoning gap in their generative formulation. This reveals that spatial connectivity and building physics must be modeled jointly, rather than treated as post-hoc adjustments.
In contrast, our GreenFlow (edge/graph) generates layouts with coherent room organization and context-aware opening allocation, enabling both functional circulation and realistic thermal behavior. The resulting designs achieve lower EUI and higher spatial rationality without requiring manual correction. These qualitative findings reinforce our core claim: explicitly incorporating energy-aware spatial semantics during generation leads to more physically grounded and practically usable architectural layouts.

\subsection{Quantitative Comparison}
\begin{table}[t]
\centering
\renewcommand{\arraystretch}{1.2}
\caption{Comparison with baselines on layout generation quality.
\textbf{Bold} indicates the best results.
\textit{*} denotes results generated \textit{without} green performance constraints.}
\label{tab:baseline}
\resizebox{\linewidth}{!}{%
\begin{tabular}{lccccc}
\toprule
\textbf{Method} & \textbf{FID}↓ & \textbf{IoU}↑ & \textbf{GED}↓ & \textbf{Rationality}↑ & \boldmath{$\Delta$}\textbf{EUI}↓ \\
\midrule
HouseGAN & 38.1 & 0.13 & -- & 0.0\% & -- \\
HouseGAN++ & 37.4 & 0.13 & 8.6 & 13\% & -- \\
HouseDiffusion & 33.2 & 0.16 & 4.2 & 24\% & -- \\
Graph2Plan & 33.7 & 0.19 & 4.8 & 22\% & +0.2\% \\
Tell2Design & 25.2 & 0.53 & -- & -- & -- \\
FloorplanDiff. & 20.2 & 0.48 & 11.7 & -- & +0.7\% \\
FP-LLaMa & 28.2 & 0.12 & -- & 38.5\% & -0.2\% \\
\midrule
\textbf{$\text{GreenFlow}_{\text{RPLAN}}^*$ (graph)} & 17.6 & 0.84 & 2.8 & 59.4\% & -2.3\% \\
\textbf{$\text{GreenFlow}_{\text{RPLAN}}$ (graph)} & 15.2 & 0.75 & 5.9 & 57.8\% & -6.9\% \\
\textbf{$\text{GreenFlow}_{\text{GreenPD}}^*$ (graph)} & 13.3 & \textbf{0.84} & \textbf{2.5} & \textbf{61.6\%} & -2.5\% \\
\textbf{$\text{GreenFlow}_{\text{GreenPD}}$ (graph)} & \textbf{13.1} & 0.75 & 5.7 & 58.8\% & \textbf{-10.4\%} \\
\midrule
\textbf{$\text{GreenFlow}_{\text{RPLAN}}^*$ (edge)} & 17.9 & 0.29 & -- & 51.3\% & -3.8\% \\
\textbf{$\text{GreenFlow}_{\text{RPLAN}}$ (edge)} & 17.4 & 0.28 & -- & 48.1\% & -12.6\% \\
\textbf{$\text{GreenFlow}_{\text{GreenPD}}^*$ (edge)} & 15.8 & \textbf{0.41} & -- & \textbf{70.6\%} & -1.9\% \\
\textbf{$\text{GreenFlow}_{\text{GreenPD}}$ (edge)} & \textbf{15.7} & 0.37 & -- & 66.3\% & \textbf{-13.1\%} \\
\bottomrule
\end{tabular}
}
\end{table}

Several baselines (e.g., HouseDiffusion, HouseGAN++) miss openings such as windows, leading to incomplete layouts, while others (e.g., Tell2Design, HouseGAN) lack essential functional components. Hence, these methods are excluded from quantitative energy evaluation.

\paragraph{Metrics Analysis.}
Tab.~\ref{tab:baseline} shows that GreenFlow consistently outperforms all baselines across both spatial and energy metrics. 
It achieves the lowest FID and the highest IoU among all methods, indicating that our generated layouts are both realistic and functional consistent. 
Under the graph setting, $\text{GreenFlow}_{\text{GreenPD}}$ attains the best layout fidelity (FID = 13.1) and spatial accuracy (IoU = 0.84), significantly surpassing FP-LLaMa (28.2 / 0.12) and HouseDiffusion (33.2 / 0.16). 
It also achieves the lowest GED (2.5) and the highest Rationality (61.6\%), confirming superior spatial-functional feasibility.  
For energy efficiency, models trained \textit{without} explicit constraints (\textit{*}) reduce mean EUI by 2.5\% (graph) and 1.9\% (edge). 
When green constraints are applied, the improvements grow substantially to 10.4\% and 13.1\%, respectively, showing that our physically guided diffusion can effectively balance spatial organization with energy objectives. 
Compared with HouseDiffusion and FP-LLaMa, GreenFlow achieves consistent gains across all indicators, regardless of input modality. 
These results highlight that integrating regulation-compliant data (GreenPD) with performance-aware generation enables our model to jointly optimize geometric fidelity, functional feasibility, and physical performance.

\paragraph{Human Expert User Study.} 
\begin{table}[t]
\centering
\setlength{\tabcolsep}{3.5pt} 
\renewcommand{\arraystretch}{1.05} 
\small 
\caption{Scores Assigned by Human Experts on Design Usability, Soundness, and Efficiency.}
\begin{tabular}{lccc}
\toprule
\textbf{Method} & \textbf{Usability}~$\uparrow$ & \textbf{Soundness}~$\uparrow$ & \textbf{Efficiency}~$\uparrow$ \\
\midrule
HouseDiffusion & 67.2 & 76.4 & -- \\
Graph2Plan & 70.4 & 77.9 & -- \\
Tell2Design & 74.3 & 72.3 & -- \\
FloorplanDiffusion & 82.8 & 75.4 & -- \\
\midrule
\textbf{GreenFlow (ours)} & \textbf{89.4} & \textbf{90.2} & \textbf{87.0\%} \\
\bottomrule
\end{tabular}
\label{tab:expert}
\end{table}
We conducted a user study with 13 experts (4 architects and 9 graduate students) to evaluate \textit{usability}, \textit{soundness}, and \textit{efficiency}.
Usability and soundness were rated on a 100-point scale, and efficiency was measured by time saved compared with manual design.
As shown in Tab.~\ref{tab:expert}, GreenFlow achieved the highest ratings (89.4 in usability and 90.2 in soundness) thanks to its built-in evaluator that removes the need for external simulations.
Manual design took 24.7 minutes on average, while GreenFlow completed the same task in only 3.2 minutes (1.7 generations, 61s each), resulting in an \textit{87.0\% improvement in efficiency}.
These findings show that GreenFlow makes design work faster, easier, and more reliable.

\subsection{Ablation Study}
\begin{figure}[t]
\centering
\includegraphics[width=1\linewidth]{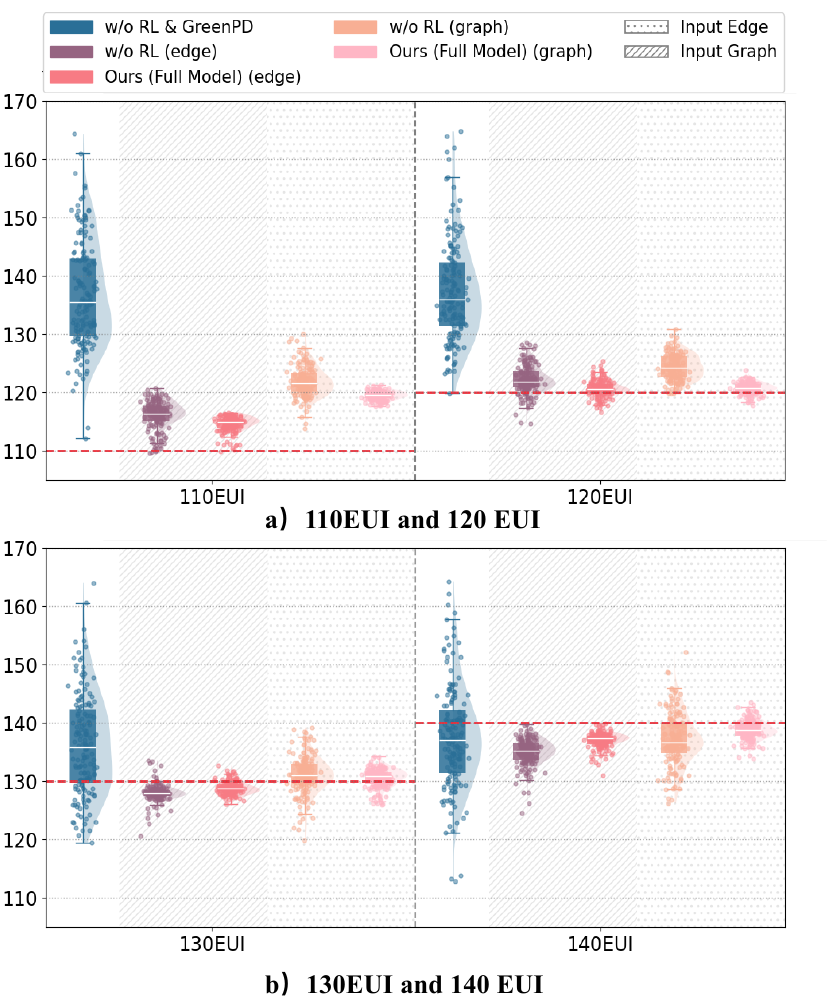}
\caption{Energy controllability under different energy targets.
Each model is evaluated at four target EUI levels (110, 120, 130, 140). 
Models trained on GreenPD (w/o RL and Full Model) are tested under both Graph and Edge inputs. 
\textit{Deviations at 110 EUI arise from functional constraints (e.g., kitchens).}}
\label{fig:energy_control}
\end{figure}

\begin{table}[t]
\centering
\small
\setlength{\tabcolsep}{3pt}
\renewcommand{\arraystretch}{1.1}
\caption{
Pass rate (\%) comparison under different input modes (\textbf{Graph} vs. \textbf{Edge}) across Base, w/o RL, and w/ RL.
}
\label{tab:ablation_passrate}
\begin{tabular}{l
                ccc
                ccc}
\toprule
\multirow{2}{*}{\textbf{Scene}} &
\multicolumn{3}{c}{\textbf{Graph}} &
\multicolumn{3}{c}{\textbf{Edge}} \\
\cmidrule(lr){2-4}\cmidrule(lr){5-7}
& \textbf{Base} & \textbf{w/o RL} & \textbf{Full} &
  \textbf{Base} & \textbf{w/o RL} & \textbf{Full} \\
\midrule
\textbf{Fire} & 41.95 & 38.75 & \textbf{79.38} & 41.95 & 84.38 & \textbf{96.25} \\
\textbf{Area} & 79.78 & 89.38 & \textbf{96.25} & 79.78 & 90.00 & \textbf{93.12} \\
\textbf{Connect} & 73.31 & 81.57 & \textbf{88.58} & 73.30 & 82.50 & \textbf{87.20} \\
\bottomrule
\end{tabular}
\end{table}

\begin{figure}[t]
\centering
\includegraphics[width=1.0\linewidth]{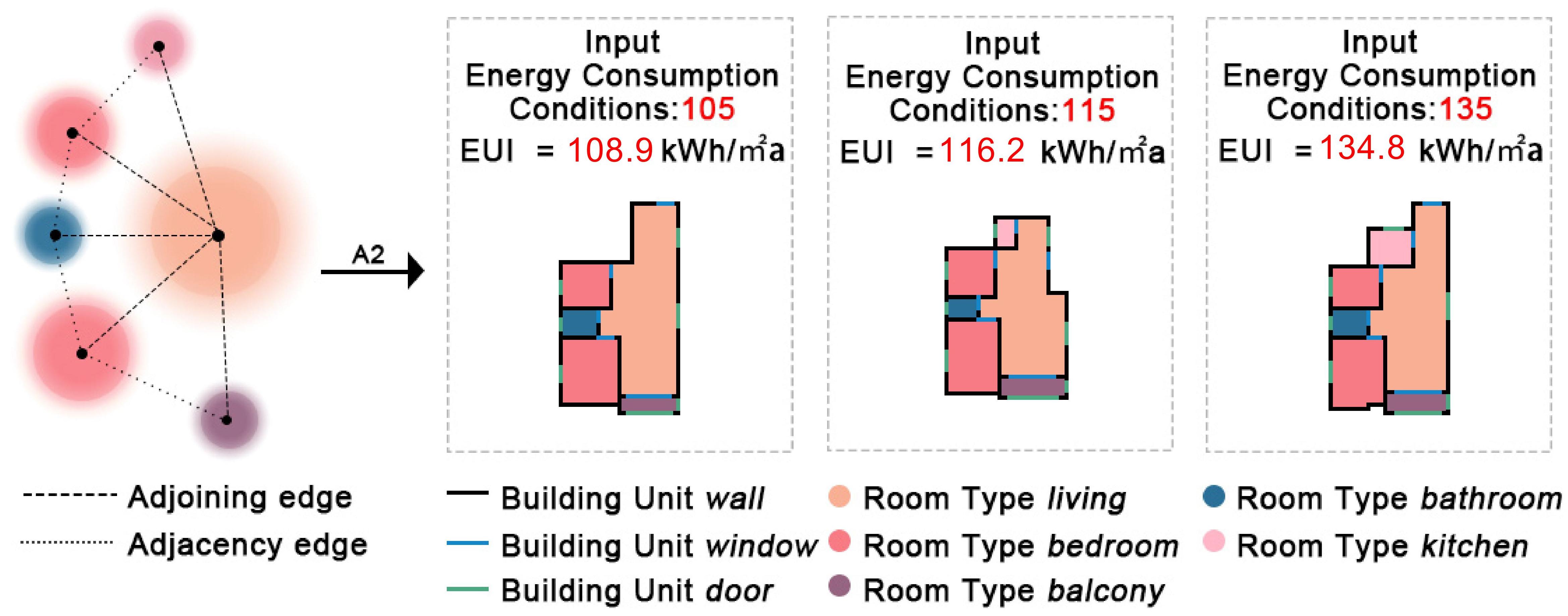}
\caption{\textbf{Controllability of green-performance attributes.}
Layouts generated under increasing target energy levels (105–135 EUI) adaptively expand high-energy functional zones (e.g., kitchens, bathrooms). 
At ultra-low targets, conflicts between functionality and energy objectives reveal the model’s adherence to both energy and functional feasibility.}
\label{fig:green_conflict}
\end{figure}

\paragraph{Ablation Settings.}
We conduct ablation studies to examine how dataset quality, RL fine-tuning, and input modality affect layout controllability:
\begin{itemize}
    \item \textbf{w/o RL \& GreenPD (Base):} trained only on the RPLAN dataset~\cite{rplan} without any energy or physical feedback, serving as a weak baseline for controllability.
    \item \textbf{w/o RL:} trained on our regulation-compliant \textit{GreenPD} dataset using both \textit{Graph} and \textit{Edge} inputs, but without PDE-based reinforcement learning. This isolates the effect of RL optimization and is evaluated under four target energy levels (110–140 EUI).
    \item \textbf{Ours (Full):} GreenFlow model trained on GreenPD and fine-tuned with PDE-guided reinforcement learning for joint optimization of energy and functional feasibility.
\end{itemize}

\paragraph{Effect of Dataset Quality.}
Replacing the noisy RPLAN dataset with our regulation-compliant GreenPD significantly improves layout stability, energy alignment, and spatial-functional feasibility. 
As shown in Tab.~\ref{tab:baseline} and Fig.~\ref{fig:energy_control}, models trained on GreenPD exhibit lower variance and stronger correspondence to target EUI values compared with those trained on RPLAN. 
This demonstrates that consistent, regulation-compliant data are essential for reliable, controllable layout generation.

\paragraph{Effect of RL Fine-tuning.}
Comparing the GreenPD-only variant (w/o RL) with the RL-enhanced model (Full) shows consistent improvements in pass rate across both input modes (Tab.~\ref{tab:ablation_passrate}). 
Notably, for \textit{Fire}, the pass rate increases from 38.75\% to 79.38\% under the \textit{Graph} setting and from 84.38\% to 96.25\% under \textit{Edge}, corresponding to gains of 40.6 and 11.9 percentage points, respectively. 
Together with the tighter EUI control shown in Fig.~\ref{fig:energy_control}, these results demonstrate that RL fine-tuning effectively enhances energy–function coupling and overall design feasibility.

\paragraph{Effect of Input Modality.}
To examine the influence of input modality, we analyze results under \textit{Graph} and \textit{Edge} settings. 
Both modalities achieve effective control, but show distinct behaviors: 
Edge inputs provide more direct control over total area and better adherence to ultra-low targets (110 EUI), while Graph inputs offer richer spatial relationships and higher functional realism. 
As visualized in Fig.~\ref{fig:green_conflict}, the model adaptively expands or suppresses high-energy functional zones (e.g., kitchens, bathrooms) as target EUI increases, maintaining a balance between physical performance and functional feasibility.

%% file: sec/5_conclusion.tex
\section{Conclusion}
We introduced GreenPlanner, a framework for generating floorplans that are both functionally feasible and energy-efficient. GreenPlanner achieves this by coupling a practical design evaluator with reinforcement-guided generation, unifying generation and feasibility assessment within a single pipeline, and allowing building regulations and energy constraints to be incorporated during generation. 

%% file: main.bib
@inproceedings{johnson2016perceptual,
  title={Perceptual losses for real-time style transfer and super-resolution},
  author={Johnson, Justin and Alahi, Alexandre and Fei-Fei, Li},
  booktitle={European conference on computer vision},
  pages={694--711},
  year={2016},
  organization={Springer}
}

@article{rplan,
  title={Data-driven interior plan generation for residential buildings},
  author={Wu, Wenming and Fu, Xiao-Ming and Tang, Rui and Wang, Yuhan and Qi, Yu-Hao and Liu, Ligang},
  journal={ACM Transactions on Graphics (TOG)},
  volume={38},
  number={6},
  pages={1--12},
  year={2019},
  publisher={ACM New York, NY, USA}
}

@inproceedings{housegan++,
  title={House-gan++: Generative adversarial layout refinement network towards intelligent computational agent for professional architects},
  author={Nauata, Nelson and Hosseini, Sepidehsadat and Chang, Kai-Hung and Chu, Hang and Cheng, Chin-Yi and Furukawa, Yasutaka},
  booktitle={CVPR},
  pages={13632--13641},
  year={2021}
}

@article{Graph2Plan,
  title={Graph2plan: Learning floorplan generation from layout graphs},
  author={Hu, Ruizhen and Huang, Zeyu and Tang, Yuhan and Van Kaick, Oliver and Zhang, Hao and Huang, Hui},
  journal={ACM Transactions on Graphics (TOG)},
  volume={39},
  number={4},
  pages={118--1},
  year={2020},
  publisher={ACM New York, NY, USA}
}

@inproceedings{HouseGAN,
  title={House-gan: Relational generative adversarial networks for graph-constrained house layout generation},
  author={Nauata, Nelson and Chang, Kai-Hung and Cheng, Chin-Yi and Mori, Greg and Furukawa, Yasutaka},
  booktitle={European Conference on Computer Vision},
  pages={162--177},
  year={2020},
  organization={Springer}
}

@inproceedings{FloorplanDiffusion,
  title={Floorplan-Diffusion: Automatic Floor Plan Generation via Pre-trained Large Latent Diffusion Model},
  author={Xu, Minyang and Lou, Yunzhong and Gao, Xiang and Zhou, Xiangdong},
  booktitle={Proceedings of the 2025 International Conference on Multimedia Retrieval},
  pages={1617--1625},
  year={2025}
}

@article{dales2008quality,
  title={Quality of indoor residential air and health},
  author={Dales, Robert and Liu, Ling and Wheeler, Amanda J and Gilbert, Nicolas L},
  journal={Cmaj},
  volume={179},
  number={2},
  pages={147--152},
  year={2008},
  publisher={CMAJ}
}

@article{baduge2022artificial,
  title={Artificial intelligence and smart vision for building and construction 4.0: Machine and deep learning methods and applications},
  author={Baduge, Shanaka Kristombu and Thilakarathna, Sadeep and Perera, Jude Shalitha and Arashpour, Mehrdad and Sharafi, Pejman and Teodosio, Bertrand and Shringi, Ankit and Mendis, Priyan},
  journal={Automation in Construction},
  volume={141},
  pages={104440},
  year={2022},
  publisher={Elsevier}
}

@article{schulman2017proximal,
  title={Proximal policy optimization algorithms},
  author={Schulman, John and Wolski, Filip and Dhariwal, Prafulla and Radford, Alec and Klimov, Oleg},
  journal={arXiv preprint arXiv:1707.06347},
  year={2017}
}

@inproceedings{clip,
  title={Learning transferable visual models from natural language supervision},
  author={Radford, Alec and Kim, Jong Wook and Hallacy, Chris and Ramesh, Aditya and Goh, Gabriel and Agarwal, Sandhini and Sastry, Girish and Askell, Amanda and Mishkin, Pamela and Clark, Jack and others},
  booktitle={ICML},
  pages={8748--8763},
  year={2021},
  organization={PmLR}
}

@article{adamw,
  title={Decoupled weight decay regularization},
  author={Loshchilov, Ilya and Hutter, Frank},
  journal={arXiv preprint arXiv:1711.05101},
  year={2017}
}

@article{everingham2010pascal,
  title={The pascal visual object classes (voc) challenge},
  author={Everingham, Mark and Van Gool, Luc and Williams, Christopher KI and Winn, John and Zisserman, Andrew},
  journal={IJCV},
  year={2010},
  publisher={Springer}
}

@article{heusel2017gans,
  title={Gans trained by a two time-scale update rule converge to a local nash equilibrium},
  author={Heusel, Martin and Ramsauer, Hubert and Unterthiner, Thomas and Nessler, Bernhard and Hochreiter, Sepp},
  journal={NeurIPS},
  year={2017}
}

@article{lehmann2024definitive,
  title={The definitive guide to policy gradients in deep reinforcement learning: Theory, algorithms and implementations},
  author={Lehmann, Matthias},
  journal={arXiv preprint arXiv:2401.13662},
  year={2024}
}

@article{vaswani2017attention,
  title={Attention is all you need},
  author={Vaswani, Ashish and Shazeer, Noam and Parmar, Niki and Uszkoreit, Jakob and Jones, Llion and Gomez, Aidan N and Kaiser, {\L}ukasz and Polosukhin, Illia},
  journal={Advances in neural information processing systems},
  volume={30},
  year={2017}
}

@inproceedings{he2016deep,
  title={Deep residual learning for image recognition},
  author={He, Kaiming and Zhang, Xiangyu and Ren, Shaoqing and Sun, Jian},
  booktitle={Proceedings of the IEEE conference on computer vision and pattern recognition},
  pages={770--778},
  year={2016}
}

@article{chang2025design,
  title={On the design fundamentals of diffusion models: A survey},
  author={Chang, Ziyi and Koulieris, George A and Chang, Hyung Jin and Shum, Hubert PH},
  journal={Pattern Recognition},
  pages={111934},
  year={2025},
  publisher={Elsevier}
}

@article{sengar2025generative,
  title={Generative artificial intelligence: a systematic review and applications},
  author={Sengar, Sandeep Singh and Hasan, Affan Bin and Kumar, Sanjay and Carroll, Fiona},
  journal={Multimedia Tools and Applications},
  volume={84},
  number={21},
  pages={23661--23700},
  year={2025},
  publisher={Springer}
}

@article{croitoru2023diffusion,
  title={Diffusion models in vision: A survey},
  author={Croitoru, Florinel-Alin and Hondru, Vlad and Ionescu, Radu Tudor and Shah, Mubarak},
  journal={IEEE transactions on pattern analysis and machine intelligence},
  volume={45},
  number={9},
  pages={10850--10869},
  year={2023},
  publisher={Ieee}
}

@article{papamakarios2021normalizing,
  title={Normalizing flows for probabilistic modeling and inference},
  author={Papamakarios, George and Nalisnick, Eric and Rezende, Danilo Jimenez and Mohamed, Shakir and Lakshminarayanan, Balaji},
  journal={Journal of Machine Learning Research},
  volume={22},
  number={57},
  pages={1--64},
  year={2021}
}

@article{kobyzev2020normalizing,
  title={Normalizing flows: An introduction and review of current methods},
  author={Kobyzev, Ivan and Prince, Simon JD and Brubaker, Marcus A},
  journal={IEEE transactions on pattern analysis and machine intelligence},
  volume={43},
  number={11},
  pages={3964--3979},
  year={2020},
  publisher={IEEE}
}

@article{li2024revision,
  title={Revision matters: Generative design guided by revision edits},
  author={Li, Tao and Cheng, Chin-Yi and Xie, Amber and Li, Gang and Li, Yang},
  journal={arXiv preprint arXiv:2406.18559},
  year={2024}
}

@article{alvur2025potential,
  title={The potential and challenges of Bim in enhancing energy efficiency in existing buildings: A comprehensive review},
  author={Alvur, Emre and Ana{\c{c}}, Merve and Cuce, Pinar Mert and Cuce, Erdem},
  journal={Sustainable and Clean Buildings},
  pages={42--65},
  year={2025}
}

@inproceedings{liu2025large,
  title={Large language models for building energy applications: Opportunities and challenges},
  author={Liu, Mingzhe and Zhang, Liang and Chen, Jianli and Chen, Wei-An and Yang, Zhiyao and Lo, L James and Wen, Jin and O’Neill, Zheng},
  booktitle={Building Simulation},
  volume={18},
  number={2},
  pages={225--234},
  year={2025},
  organization={Springer}
}

@article{ding2025understanding,
  title={Understanding world or predicting future? a comprehensive survey of world models},
  author={Ding, Jingtao and Zhang, Yunke and Shang, Yu and Zhang, Yuheng and Zong, Zefang and Feng, Jie and Yuan, Yuan and Su, Hongyuan and Li, Nian and Sukiennik, Nicholas and others},
  journal={ACM Computing Surveys},
  year={2025},
  publisher={ACM New York, NY}
}

@article{hu2025simulating,
  title={Simulating the real world: A unified survey of multimodal generative models},
  author={Hu, Yuqi and Wang, Longguang and Liu, Xian and Chen, Ling-Hao and Guo, Yuwei and Shi, Yukai and Liu, Ce and Rao, Anyi and Wang, Zeyu and Xiong, Hui},
  journal={arXiv preprint arXiv:2503.04641},
  year={2025}
}

@article{kingma2013auto,
  title={Auto-encoding variational bayes},
  author={Kingma, Diederik P},
  journal={arXiv preprint arXiv:1312.6114},
  year={2013}
}

@article{goodfellow2020generative,
  title={Generative adversarial networks},
  author={Goodfellow, Ian and Pouget-Abadie, Jean and Mirza, Mehdi and Xu, Bing and Warde-Farley, David and Ozair, Sherjil and Courville, Aaron and Bengio, Yoshua},
  journal={Communications of the ACM},
  volume={63},
  year={2020},
  publisher={ACM New York, NY, USA}
}

@article{ho2020denoising,
  title={Denoising diffusion probabilistic models},
  author={Ho, Jonathan and Jain, Ajay and Abbeel, Pieter},
  journal={Advances in neural information processing systems},
  volume={33},
  pages={6840--6851},
  year={2020}
}

@article{song2020denoising,
  title={Denoising diffusion implicit models},
  author={Song, Jiaming and Meng, Chenlin and Ermon, Stefano},
  journal={arXiv preprint arXiv:2010.02502},
  year={2020}
}

@article{bond2021deep,
  title={Deep generative modelling: A comparative review of vaes, gans, normalizing flows, energy-based and autoregressive models},
  author={Bond-Taylor, Sam and Leach, Adam and Long, Yang and Willcocks, Chris G},
  journal={IEEE transactions on pattern analysis and machine intelligence},
  volume={44},
  number={11},
  pages={7327--7347},
  year={2021},
  publisher={IEEE}
}

@article{lee2023simplified,
  title={Simplified methods for generative design that combine evaluation techniques for automated conceptual building design},
  author={Lee, Jaewook and Cho, Wonho and Kang, Dongyeop and Lee, Jongho},
  journal={Applied Sciences},
  volume={13},
  year={2023},
  publisher={MDPI}
}

@article{regenwetter2023beyond,
  title={Beyond statistical similarity: Rethinking metrics for deep generative models in engineering design},
  author={Regenwetter, Lyle and Srivastava, Akash and Gutfreund, Dan and Ahmed, Faez},
  journal={Computer-Aided Design},
  volume={165},
  pages={103609},
  year={2023},
  publisher={Elsevier}
}

@article{chen2025gdt,
  title={GDT framework: integrating generative design and design thinking for sustainable development in the AI era},
  author={Chen, Yongliang and Qin, Zhongzhi and Sun, Li and Wu, Jiantao and Ai, Wen and Chao, Jiayuan and Li, Huaixin and Li, Jiangnan},
  journal={Sustainability},
  volume={17},
  pages={372},
  year={2025},
  publisher={Multidisciplinary Digital Publishing Institute}
}

@inproceedings{holodeck,
  title={Holodeck: Language guided generation of 3d embodied ai environments},
  author={Yang, Yue and Sun, Fan-Yun and Weihs, Luca and VanderBilt, Eli and Herrasti, Alvaro and Han, Winson and Wu, Jiajun and Haber, Nick and Krishna, Ranjay and Liu, Lingjie and others},
  booktitle={CVPR},
  pages={16227--16237},
  year={2024}
}

@article{layoutgpt,
  title={Layoutgpt: Compositional visual planning and generation with large language models},
  author={Feng, Weixi and Zhu, Wanrong and Fu, Tsu-jui and Jampani, Varun and Akula, Arjun and He, Xuehai and Basu, Sugato and Wang, Xin Eric and Wang, William Yang},
  journal={NeurIPS},
  year={2023}
}

@article{eberhardt2022building,
  title={Building design and construction strategies for a circular economy},
  author={Eberhardt, Leonora Charlotte Malabi and Birkved, Morten and Birgisdottir, Harpa},
  journal={Architectural Engineering and Design Management},
  volume={18},
  number={2},
  pages={93--113},
  year={2022},
  publisher={Taylor \& Francis}
}

@article{himeur2023ai,
  title={AI-big data analytics for building automation and management systems: a survey, actual challenges and future perspectives},
  author={Himeur, Yassine and Elnour, Mariam and Fadli, Fodil and Meskin, Nader and Petri, Ioan and Rezgui, Yacine and Bensaali, Faycal and Amira, Abbes},
  journal={Artificial intelligence review},
  volume={56},
  number={6},
  pages={4929--5021},
  year={2023},
  publisher={Springer}
}

@article{tell2design,
  title={Tell2design: A dataset for language-guided floor plan generation},
  author={Leng, Sicong and Zhou, Yang and Dupty, Mohammed Haroon and Lee, Wee Sun and Joyce, Sam Conrad and Lu, Wei},
  journal={arXiv preprint arXiv:2311.15941},
  year={2023}
}

@article{chen2025comprehensive,
  title={Comprehensive exploration of diffusion models in image generation: a survey},
  author={Chen, Hang and Xiang, Qian and Hu, Jiaxin and Ye, Meilin and Yu, Chao and Cheng, Hao and Zhang, Lei},
  journal={Artificial Intelligence Review},
  year={2025},
  publisher={Springer}
}

@article{li2025comprehensive,
  title={A comprehensive survey of image generation models based on deep learning},
  author={Li, Jun and Zhang, Chenyang and Zhu, Wei and Ren, Yawei},
  journal={Annals of Data Science},
  year={2025},
}

@article{wang2025diffusion,
  title={Diffusion-based visual art creation: A survey and new perspectives},
  author={Wang, Bingyuan and Chen, Qifeng and Wang, Zeyu},
  journal={ACM Computing Surveys},
  year={2025},
}

@article{xue2025human,
  title={Human motion video generation: A survey},
  author={Xue, Haiwei and Luo, Xiangyang and Hu, Zhanghao and Zhang, Xin and Xiang, Xunzhi and Dai, Yuqin and Liu, Jianzhuang and Zhang, Zhensong and Li, Minglei and Yang, Jian and others},
  journal={TPAMI},
  year={2025},
  publisher={IEEE}
}

@article{gemini,
  title={Gemini 2.5: Pushing the frontier with advanced reasoning, multimodality, long context, and next generation agentic capabilities},
  author={Comanici, Gheorghe and Bieber, Eric and Schaekermann, Mike and Pasupat, Ice and Sachdeva, Noveen and Dhillon, Inderjit and Blistein, Marcel and Ram, Ori and Zhang, Dan and Rosen, Evan and others},
  journal={arXiv preprint arXiv:2507.06261},
  year={2025}
}

@article{gpt4o,
  title={Gpt-4o system card},
  author={Hurst, Aaron and Lerer, Adam and Goucher, Adam P and Perelman, Adam and Ramesh, Aditya and Clark, Aidan and Ostrow, AJ and Welihinda, Akila and Hayes, Alan and Radford, Alec and others},
  journal={arXiv preprint arXiv:2410.21276},
  year={2024}
}

@inproceedings{housediffusion,
  title={Housediffusion: Vector floorplan generation via a diffusion model with discrete and continuous denoising},
  author={Shabani, Mohammad Amin and Hosseini, Sepidehsadat and Furukawa, Yasutaka},
  booktitle={CVPR},
  year={2023}
}

@article{zhang2023leveraging,
  title={Leveraging policy instruments and financial incentives to reduce embodied carbon in energy retrofits},
  author={Zhang, Haonan},
  journal={arXiv preprint arXiv:2304.03403},
  year={2023}
}

@article{zhang2024globus,
  title={GLOBUS: Global building renovation potential by 2070},
  author={Zhang, Shufan and Ma, Minda and Zhou, Nan and Yan, Jinyue},
  journal={arXiv preprint arXiv:2406.04133},
  year={2024}
}

@article{liang2023decarbonization,
  title={Decarbonization potentials of the embodied energy use and operational process in buildings: A review from the life-cycle perspective},
  author={Liang, Yumin and Li, Changqi and Liu, Zhichao and Wang, Xi and Zeng, Fei and Yuan, Xiaolei and Pan, Yiqun},
  journal={Heliyon},
  volume={9},
  number={10},
  year={2023},
  publisher={Elsevier}
}

@inproceedings{yin2025floorplan,
  title={Floorplan-llama: Aligning architects’ feedback and domain knowledge in architectural floor plan generation},
  author={Yin, Jun and Zeng, Pengyu and Sun, Haoyuan and Dai, Yuqin and Zheng, Han and Zhang, Miao and Zhang, Yachao and Lu, Shuai},
  booktitle={Proceedings of the 63rd Annual Meeting of the Association for Computational Linguistics (Volume 1: Long Papers)},
  pages={6640--6662},
  year={2025}
}

@article{zeng2024residential,
  title={Residential floor plans: Multi-conditional automatic generation using diffusion models},
  author={Zeng, Pengyu and Gao, Wen and Yin, Jun and Xu, Pengjian and Lu, Shuai},
  journal={Automation in Construction},
  volume={162},
  pages={105374},
  year={2024},
  publisher={Elsevier}
}

@article{zeng2025automated,
  title={Automated residential layout generation and editing using natural language and images},
  author={Zeng, Pengyu and Gao, Wen and Li, Jizhizi and Yin, Jun and Chen, Jiling and Lu, Shuai},
  journal={Automation in Construction},
  volume={174},
  pages={106133},
  year={2025},
  publisher={Elsevier}
}

@inproceedings{zeng2025card,
  title={Card: Cross-modal agent framework for generative and editable residential design},
  author={Zeng, Pengyu and Yin, Jun and Zhang, Miao and Dai, Yuqin and Li, Jizhizi and Jin, Zhanxiang and Lu, Shuai},
  booktitle={Proceedings of the 2025 Conference on Empirical Methods in Natural Language Processing},
  pages={9315--9330},
  year={2025}
}

@article{zeng2025comprehensive,
  title={Comprehensive and dedicated metrics for evaluating AI-generated residential floor plans},
  author={Zeng, Pengyu and Yin, Jun and Gao, Yan and Li, Jizhizi and Jin, Zhanxiang and Lu, Shuai},
  journal={Buildings},
  volume={15},
  number={10},
  pages={1674},
  year={2025},
  publisher={MDPI}
}

@article{zeng2025unified,
  title={Unified residential floor plan generation with multimodal inputs},
  author={Zeng, Pengyu and Yin, Jun and Zhang, Miao and Li, Jizhizi and Zhang, Yachao and Lu, Shuai},
  journal={Automation in Construction},
  volume={178},
  pages={106408},
  year={2025},
  publisher={Elsevier}
}

@article{yin2025floorplanmae,
  title={FloorplanMAE: A self-supervised framework for complete floorplan generation from partial inputs},
  author={Yin, Jun and Zhong, Jing and Zeng, Pengyu and Li, Peilin and Zhang, Miao and Luo, Ran and Lu, Shuai},
  journal={arXiv preprint arXiv:2506.08363},
  year={2025}
}

@article{yin2025drag2build++,
  title={Drag2Build++: A drag-based 3D architectural mesh editing workflow based on differentiable surface modeling},
  author={Yin, Jun and Zeng, Pengyu and Li, Peilin and Zhong, Jing and Hao, Tianze and Zheng, Han and Lu, Shuai},
  journal={Frontiers of Architectural Research},
  year={2025},
  publisher={Elsevier}
}

@inproceedings{yin2025archiset,
  title={ArchiSet: Benchmarking Editable and Consistent Single-View 3D Reconstruction of Buildings with Specific Window-to-Wall Ratios},
  author={Yin, Jun and Zeng, Pengyu and Shen, Licheng and Zhang, Miao and Zhong, Jing and Han, Yuxing and Lu, Shuai},
  booktitle={Proceedings of the IEEE/CVF International Conference on Computer Vision},
  pages={26004--26014},
  year={2025}
}

@inproceedings{zeng2025ai,
  title={AI-based generation and optimization of energy-efficient residential layouts controlled by contour and room number},
  author={Zeng, Pengyu and Yin, Jun and Huang, Yujian and Zhong, Jing and Lu, Shuai},
  booktitle={Building Simulation},
  volume={18},
  number={10},
  pages={2777--2805},
  year={2025},
  organization={Springer}
}

@inproceedings{zeng2025mred,
  title={MRED-14: A Benchmark for Low-Energy Residential Floor Plan Generation with 14 Flexible Inputs},
  author={Zeng, Pengyu and Yin, Jun and Sun, Haoyuan and Dai, Yuqin and Jiang, Maowei and Zhang, Miao and Lu, Shuai},
  booktitle={Proceedings of the 33rd ACM International Conference on Multimedia},
  pages={11298--11307},
  year={2025}
}

@article{jiang2023fecam,
  title={FECAM: Frequency enhanced channel attention mechanism for time series forecasting},
  author={Jiang, Maowei and Zeng, Pengyu and Wang, Kai and Liu, Huan and Chen, Wenbo and Liu, Haoran},
  journal={Advanced Engineering Informatics},
  volume={58},
  pages={102158},
  year={2023},
  publisher={Elsevier}
}

@article{zeng2022muformer,
  title={Muformer: A long sequence time-series forecasting model based on modified multi-head attention},
  author={Zeng, Pengyu and Hu, Guoliang and Zhou, Xiaofeng and Li, Shuai and Liu, Pengjie and Liu, Shurui},
  journal={Knowledge-Based Systems},
  volume={254},
  pages={109584},
  year={2022},
  publisher={Elsevier}
}

@article{zeng2023seformer,
  title={Seformer: a long sequence time-series forecasting model based on binary position encoding and information transfer regularization},
  author={Zeng, Pengyu and Hu, Guoliang and Zhou, Xiaofeng and Li, Shuai and Liu, Pengjie},
  journal={Applied Intelligence},
  volume={53},
  number={12},
  pages={15747--15771},
  year={2023},
  publisher={Springer}
}

@inproceedings{dai2025harmonious,
  title={Harmonious Music-driven Group Choreography with Trajectory-Controllable Diffusion},
  author={Dai, Yuqin and Zhu, Wanlu and Li, Ronghui and Ren, Zeping and Zhou, Xiangzheng and Ying, Jixuan and Li, Jun and Yang, Jian},
  booktitle={Proceedings of the AAAI Conference on Artificial Intelligence},
  volume={39},
  number={3},
  pages={2645--2653},
  year={2025}
}

@article{dai2025mindaligner,
  title={MindAligner: Explicit Brain Functional Alignment for Cross-Subject Visual Decoding from Limited fMRI Data},
  author={Dai, Yuqin and Yao, Zhouheng and Song, Chunfeng and Zheng, Qihao and Mai, Weijian and Peng, Kunyu and Lu, Shuai and Ouyang, Wanli and Yang, Jian and Wu, Jiamin},
  journal={arXiv preprint arXiv:2502.05034},
  year={2025}
}

@article{dai2025tcdiff++,
  title={TCDiff++: An End-to-end Trajectory-Controllable Diffusion Model for Harmonious Music-Driven Group Choreography},
  author={Dai, Yuqin and Zhu, Wanlu and Li, Ronghui and Li, Xiu and Zhang, Zhenyu and Li, Jun and Yang, Jian},
  journal={arXiv preprint arXiv:2506.18671},
  year={2025}
}
